# Multi-robot coordination for connectivity recovery after unpredictable environment changes

**Yaroslav Marchukov and Luis Montano** *

* *Instituto de Investigación en Ingeniería de Aragon (I3A)*
*University of Zaragoza, Spain*
*(e-mail: yamar@unizar.es, montano@unizar.es)*

**Abstract:**
In the present paper we develop a distributed method to reconnect a multi-robot team after connectivity failures, caused by unpredictable environment changes, i.e. appearance of new obstacles. After the changes, the team is divided into different groups of robots. The groups have a limited communication range and only a partial information in their field of view about the current scenario. Their objective is to form a chain from a static base station to a goal location. In the proposed distributed replanning approach, the robots predict new plans for the other groups from the new observed information by each robot in the changed scenario, to restore the connectivity with a base station and reach the initial joint objective. If a solution exists, the method achieves the reconnection of all the groups in a unique chain. The proposed method is compared with other two cases: 1) when all the agents have full information of the environment, and 2) when some robots must move to reach other waiting robots for reconnection. Numerical simulations are provided to evaluate the proposed approach in the presence of unpredictable scenario changes.

## 1. INTRODUCTION

The communication in multi-robot teams is crucial, for exchange of data between the agents during their mission. The communication between them may be permanent or intermittent, dependant on the scenario and/or application. A lot of work has been done in these fields. In SLAM applications Burgard et al. (2005), the agents periodically meet each other to share the mapped segments and reduce the localization error. In critical scenarios, such as exploration missions in disaster or robotized intervention scenarios, it is safer to keep permanent connectivity between the agents, Tardioli et al. (2016).

However, all these solutions ignore the failures that may be produced in the communication during the execution of the mission. In this work, we consider the appearance of sporadic and unpredictable events during the mission that alter the environment and therefore, produce failures in communication, i.e. disconnection of robots in the team. We contemplate common and also critical situations for a robot, such as opening and closure of doors, appearance of new obstacles. All these events break the communication between the agents, and the team is split into different groups. So that each group may be formed by a single or several agents. This fact makes a centralized solution unfeasible, because the communication between them is broken.

The contribution of this work is a distributed algorithm to recover the connectivity among the team in order to fulfill the mission. We focus here on the case where a robot has to reach a goal and the other robots act as relays forming a chain to maintain the connectivity with the base

* This research has been funded by project DPI2016-76676-R-AEI/FEDER-UE and by research grant BES-2013-067405 of MINECO-FEDER, and by project Grupo DGA-T45-17R/FSE

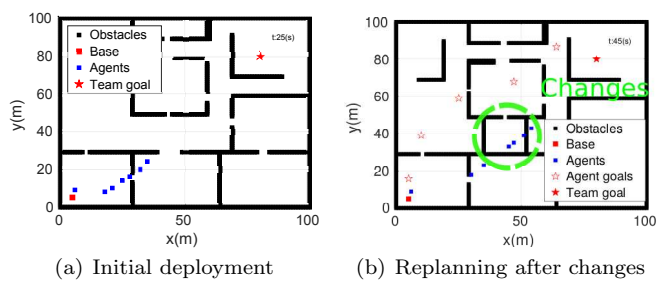

(a) Initial deployment  (b) Replanning after changes

Fig. 1. Team deployment. In (b), the 3 groups replan the objectives after the changes.

station, in a multi-hop data transmission. An illustrated example is depicted in Fig.1. The agents dispose of an initial map of the scenario, and obtain a plan to reach an objective, starting the deployment mission, Fig.1(a). As it is common in real world, the environment may change. Some doors can be opened, others to be closed or new obstacles can suddenly appear, Fig.1(b). In this case, the team of agents is divided into three groups, do not having communication with each other. Every group of agents predicts the actions of the other groups based on the knowledge of their previous information, which may also have changed. They find out a new solution, if it exists, to simultaneously recover the communication and fulfill the mission of forming a chain. The new agent goals, in the chain, are depicted with red stars in Fig.1(b). The algorithm is decentralized from the point of view of the agents in the same group. Each agent can execute the algorithm, because all of them own the same information. The algorithm is distributed from the group perspective, since the information of each of them is different.

The rest of the paper is organized as follows. In Sect.2 we present some related works. In Sect.3 we describe the problem of making predictions without full information of the entire environment. Sect.4 presents the proposed distributed algorithm. In Sect.4.1, we describe the chain formation and in Sect.4.2 we develop our prediction based planner. In Sect.5 we validate the proposed approach by means of simulations of random variations of the scenario, and discuss the results. And finally, the conclusions and future work are presented in Sect.6.

## 2. RELATED WORK

In our previous work Marchukov and Montano (2017), we developed a communication-aware deployment planer for a multi-roboIn this work, w team. In this work, when the team is executing an exploration mission and detects alterations of the scenario, it relaunches the planner with new environment information, avoiding the interruption of connectivity. This way, the agents never lose communication with the rest of the team. In the present work we consider the situation in which the communication between the team is interrupted, and the agents must restore the connectivity after appearance of new obstacles, in order to accomplish their mission. In Ghedini et al. (2018), the authors deal with stochastic failures of the nodes by means of a control law that increases the connectivity of each agent in the graph of agents. This method also anticipates to the total failure of communication in the net of agents, but not after the failure as in the present work. Furthermore, the considered failures are tied to a total malfunction of some agents, but not to failures produced due to the environment changes, as the ones considered in this paper.

When a group of agents is disconnected from others, it does not have any information about them. So, it has two ways of actuation: either to establish some kind of policy when disconnection occurs, so that, some agent will search for the rest to reconnect the entire team; or to predict the possible actions of the rest of the groups, replanning the mission to simultaneously restore the connectivity. In this paper we develop a distributed planner which uses predictions about the other groups, comparing it with a reconnection policy adapted to our problem, described in Sect.5.

An example of policy establishment is proposed in Sujit and Sousa (2012), where a team of UAVs uses a bidding policy to allocate goals for a team of robots with communication faults. The authors of Ulam and Arkin (2004) study different behaviours for the robots to recover the communication, subject to some specific locations of the scenario. We consider impossible to properly choose a secure location where the connectivity can be restored when unpredictable alterations of the environment happen. The authors of Habibi and McLurkin (2014) and Habibi et al. (2016) develop distributed methods to efficiently recover lost members. Here the connected agents form a tree, acting as relays, to guide lost robots to a goal location with connectivity. In our method, all the robots are actively looking for recover connectivity by means of predictions about the plans of other groups, do not pre-defining specific locations to restore the connection.

The prediction of the actions of the robots implies planning under uncertainty of the actions of the other agents. Several works have dealt with planning under uncertainty, as the belief roadmaps (BRM), Prentice and Roy (2009), or as the belief rapidly exploring random trees (BRRT), Bry and Roy (2011). Both methods include the uncertainties of the positions of the agents over the paths, particularly useful for problems such as SLAM, analyzed in both articles. In Indelman (2018), a probabilistic roadmap (PRM) is developed to obtain paths for the agents of a multi-robot team where they are able to reduce their localization uncertainty. In Amato et al. (2016), the authors apply a POMDP technique to coordinate a team of bartender-waiters robots, using a probabilistic model for the uncertainty of the sensing devices.

In all the aforementioned methods the uncertainty is associated to the localization error, so that can be modelled. However, in our particular problem, the appearance of the obstacles is sporadic and unpredictable, thus no uncertainty models can be used. We propose a solution where the disconnected groups predict the behavior of the agents of the other groups, based on the information available at the instant of the disconnection. Thus, a policy to reconnect the groups in a chain from the base station is applied.

## 3. PROBLEM DEFINITION

Let us denote $A = \{a_1, ..., a_N\}$ as a set of $N$ agents, which must reach a goal location and execute some task. The task is indicated from a static base station. So the agents must form a multi-hop network from the base $x_{base}$ to the goal $x_{goal}$, also known as a chain. One robot reaches $x_{goal}$ and the rest act as relays. The agents plan the deployment using the initial map of the environment and start the deployment. At some moment of the mission the environment changes, causing disconnections between the agents of the team, Fig.1. So, the team is divided into different groups, expressed as $g_i$, that is $\mathbf{g} = \{g_1, .., g_M\}$, where $M$ denotes the number of groups.

Each group is formed by one or several connected agents. We assume that every agent is equipped with a wireless antenna and the communication between two agents can be established when they are within the communication range ($c_{range}$) of each other and there is line-of-sight between them. Moreover, the robots have a limited range of field of view, $v_{range}$, thus only can observe changes within it. Each group of agents share information between them. We denote the information of a group $g_i$ as $I_i$, which is basically the updated map of the environment observed by all the agents of the group. When the changes in the environment cause breakdown of communication, every group must compute a new plan to re-connect a chain. The plan of a group $g_i$ is $\pi(g_i|I_i) = \{\rho_1, \rho_2, ...\rho_N\}$, where $\rho$ stands for the path of the agent from its current location to its corresponding assigned position in the chain. This plan computes the paths for the agents of the group and the predicted paths for the agents of the rest of groups.

Each group obtains its own plan taking into account the information $I_i$, and the predicted plans of the other groups. The plan for $g_i$ can be formally expressed as:

$$\pi(g_i) = \pi\left(g_i|I_i, \pi(g_j|I_j^i)\right), j = 1, .., M, i \neq j \qquad (1)$$

$I_j^i$ denotes the information that $g_i$ has about $g_j$, and $\pi(g_j|I_j^i)$ is the plan of group $g_j$ predicted by $g_i$, computed from the new observed scenario.

## 4. PREDICTION-BASED DISTRIBUTED COORDINATION

In the statement of the problem we have established that the robots have to form a chain to connect the goal to the base station. In 4.1 we describe the algorithm that each group executes for the chain formation. In 4.2 we explain how each group computes its own plan jointly to the plans predicted for the other groups, according to eq. 1.

*4.1 Chain formation*

Alg.1 develops the procedure used by each group to compute the chain.

**Algorithm 1** Chain planner of group $g_k$

**Require:** Agent locations $\mathbf{x}_a$, Base station $x_{base}$, Team goal $x_{goal}$, Information $I$
1: $\rho_{chain} \leftarrow compute\_chain\_path(x_{base}, x_{goal}, I)$
2: $\mathbf{x}_{lg} \leftarrow compute\_local\_goals(\rho_{chain}, I, c_{range})$
3: $\langle \mathbf{x}_a, \mathbf{x}_{lg} \rangle \leftarrow hungarian(\mathbf{x}_a, \mathbf{x}_{lg}, I)$
4: $\pi(g_k) \leftarrow compute\_paths(\langle \mathbf{x}_a, \mathbf{x}_{lg} \rangle, I)$
5: **return** $\pi(g_k)$

At first, the algorithm computes the shortest path from the base to the goal, expressed as $\rho_{chain}$, using $compute\_chain\_path$ function in l.1. This way, the algorithm obtains the possible locations where to place the agents as relays. A Fast Marching Method(FMM) (Sethian (1996)) is used for the path computation. It consists in propagating a wavefront from an initial position, base station in our case, computing the distance gradient to each position of the scenario and avoiding the obstacles. Descending the gradient from $x_{goal}$ to $x_{base}$, we obtain the path $\rho_{chain}$.

All the positions where the agents will act as relays lie on this path. The algorithm distributes the relay goals with an interval of the communication range of the agents ($c_{range}$), as well as where the agents have line-of-sight between them, with $compute\_local\_goals$ function in l.2. Note that, $\mathbf{x}_{lg}$ represent the local goals for the team. Where one goal is the goal of the mission $x_{goal}$ and the rest are relay goals.

The algorithm allocates the computed goals to each agent, with $hungarian$ function in l.3. Again the FMM is used to compute the cost to reach each goal. A unique gradient computation from an agent position provides the distances to all the goal locations. Therefore, FMM is executed other $N$ times, one per agent, obtaining all the costs to the goals. The goals are allocated to the agents, obtaining tuples of agent-goal $\langle \mathbf{x}_a, \mathbf{x}_{lg} \rangle$, using the Hungarian algorithm. With this method, the sum of the distances that travel the entire team is the minimal. If the number of relay goals is lower than $N$, the Hungarian method excludes the agents which are not used the chain. Finally, the paths are obtained for every agent of the groups, descending the gradient computed by FMM, l.4. $compute\_paths$ function obtains the paths for all the agents, also using FMM.

Note that the chain cannot be deployed if: i) a path $\rho_{chain}$ does not exist from the base station to the goal; ii) there is not enough agents to place them over this path and reach the base, produced when $length(\rho_{chain}) > N * c_{range}$; iii) the obstacles distribution obstruct the line-of-sight between the relays over $\rho_{chain}$. In these cases, the algorithm will return that no solution exists and its reason.

### 4.2 Prediction-based Planner

The predictions that each group must make about the other groups, denoted in expression (1), may become intractable if the team is composed by many agents and the number of groups is large. Every group $g_i \in \mathbf{g}$ recursively predicts the plans of the rest of the groups $g_j \in \mathbf{g}, i \neq j$, whose plans are also based in predictions. According to eq.(1), the number of predictions per group would be $M^{M-1}$. However, the complexity of the prediction is significantly reduced if we do not consider all the groups, but only the relevant ones for each of them, developed in Alg.2.

The first action of the group is to update the environment information, $update\_information$ function in l.1. Here, the subindex $k$ stands for the group which is planning. The update refers to the new observed environment information from the agent sensors within the visual range $v_{range}$.

In a chain formation, each agent depends on a unique agent and only one agent depends on him, its parent and descendant, respectively. This concept can be also applied to the groups. Therefore, the algorithm obtains the groups with its own information in l.2, where the groups are formed by connected agents. Then sorts the groups based on the proximity to the base station in l.3. This way, each group depends on the previous and the next group in the chain. The number of predictions that makes every group is reduced to $2*(M-2)+2$, that is $2*(M-2)$ predictions for intermediate groups and 2 predictions for the extreme groups of the chain.

**Algorithm 2** Group planner of $g_k$

**Require:** Agent locations $\mathbf{x}_a$, Base station $x_{base}$, Team goal $x_{goal}$, Information $I_k$
1: $I_k \leftarrow update\_information(\mathbf{x}_{a_k}, I_k, v_{range})$
2: $\mathbf{g} \leftarrow obtain\_groups(\mathbf{x}_a, I_k, c_{range})$
3: $\mathbf{g} \leftarrow sort\_groups(\mathbf{x}_{base}, I_k)$
4: **for each** $g_i \in \mathbf{g}$ **do**  ▷ $i = \{1, ..., M\}, i \neq k$
5:   $I_i^k \leftarrow update\_information(\mathbf{x}_{a_k}, I_i^k, v_{range})$
    ▷ $\mathbf{x}_{a_i}$ are locations of $g_i$
6:   $\pi(g_i) \leftarrow chain\_plan(\mathbf{x}_a, x_{base}, x_{goal}, I_i^k)$  ▷ Alg. 1
7: **end for**
8: $\pi(g_k) \leftarrow prediction\_plan(\mathbf{x}_a, \pi(g_i), I_i^k)$  ▷ Alg.3
9: **return** $\pi(g_k)$

Then, the algorithm computes the plans for all the groups, l.4-7, using the chain planner (Alg. 1) and the information that the group believes that has every group $g_i$. At this point, all the plans of the rest of the groups are obtained. The algorithm predicts the individual path of each agent of each group, l.8, described in Alg.3.

**Algorithm 3** Agent prediction planner

**Require:** Agents $\mathbf{x}_a$, Plans $\pi(g_i)$, Information $I_i^k$
1: **for** $n = 1 : N$ **do**  ▷ N agents
2:   **if** $a_{n-1} \notin g(n)$ **then**  ▷ $g(n)$ is the group of agent $a_n$
3:     **if** $goal(a_{n-1}|I_{g(n-1)}^k) != goal(a_{n-1}|I_{g(n)}^k)$ **then**
4:       $\rho_n^- \leftarrow intercept(\rho_{n-1}, x_{a_n}, I_{g(n)}^k)$  ▷ $\rho_n$: path of $a_n$
5:     **end if**
6:   **end if**
7:   **if** $a_{n+1} \notin g(a_n)$ **then**
8:     **if** $goal(a_{n+1}|I_{g(n+1)}^k) != goal(a_{n+1}|I_{g(n)}^k)$ **then**
9:       $\rho_n^+ \leftarrow intercept(\rho_{n+1}, x_{a_n}, I_{g(n)}^k)$
10:    **end if**
11:  **end if**
12:  $\pi(g_k) \leftarrow \rho_n^- + \rho_n^+ + compute\_path(x_{a_n}, x_{lg_n}, I_{g(n)}^k)$
    ▷ Complete path of agent $n$
13: **end for**
14: **return** $\pi(g_k)$

The Agent planner, in Alg.3, computes the paths for all the agents $a_n \in A$ of the team, l.1. Since the groups are sorted in terms of proximity to the base station, all the agents are also sorted in the same way. So, each agent must check how acts its link in the chain that pertains to another group. First, it checks the goal of its parent in the chain, $a_{n-1}$ l.2. If the goal of the parent does not coincide according to the information of both groups, $I_{g(n-1)}^k$ and $I_{g(n)}^k$ l.3, the agent must intercept the parent to transmit the new information, l.4. This is because the parent does not have the correct goal and it is going to some position to provide connectivity to no one. This way, the group is able to re-plan its chain to connect both groups. $intercept(\rho, x, I)$ function computes a path to the closest position of the path $\rho$ from the position $x$ using information $I$.

An illustrative example of this procedure is depicted in Fig.2. The team starts the deployment with the initial information, Fig.2(a). The appearance of new obstacles divides the team into two disconnected groups, Fig.2(b). Each group observes different environment variations, depending on its visual range $v_{range}$, Fig.2(c) and 2(f). Therefore, the groups have different initial plans, obtained with Alg.1, Fig.2(d) and 2(g). In Fig.2(e), the plan of $g_1$, $\pi(g_1|I_1)$, matches the plan predicted by $g_1$ for $g_2$, $\pi(g_2|I_2^1)$, l.4-7 of Alg.2. Thus it believes that its plan is correct and it will execute it. However, the prediction $\pi(g_1|I_1^2)$ in Fig.2(h), does not match with its plan $\pi(g_2|I_2)$, l.3 in Alg.3. Therefore, $g_2$ intercepts the agent of $g_1$, Fig.2(i), as indicated in l.4 of Alg.3. After interception, the entire team is a unique group and the agents go to their respective goals, in order to form the chain, Fig.2(j). In general, if agent $a_n$ is able to intercept to $a_{n-1}$ at some point on its path $\rho_{n-1}$, it goes to that location. If not, $a_{n-1}$ will be at the end of the path $\rho_{n-1}$. Then $a_n$ transmits to $a_{n-1}$ the new information about the changed environment. The same procedure is repeated for agent whom $a_n$ provides connectivity, the descendant agent $a_{n+1}$ in the chain, l.7-11. After inform the agents of other groups, $a_{n-1}$ and $a_{n+1}$, the agent $a_n$ finds the path to its own objective in the chain, l.12. Note that the paths of all the agents which pertain to $g_k$ are the real paths that travel the agents of the group, and the rest of the paths are just predictions for agents of other groups.

When some agent attempts to access to some room, with a unique possible access where could be some teammate, and this access is blocked by a new obstacle, the robot executes a *trapped* routine. It consists in following all the walls of the room, trying to find a new possible access. This exploration procedure is necessary because the robots must ensure if it is possible or not to access to this part of the scenario. If it is not possible, the agents inside cannot be re-connected to the whole team.

Extending recursively the method to all the groups in **g** existing in a given moment, it converges to a chain solution connecting in a unique group all the robots of the team from the base station to the final goal, in the case that a solution exists. An obvious sufficient condition is that the frequency of changes in the environment is lower than the time required to find and execute the solution. Another sufficient condition is that at least one group will observe sometime the correct information of the whole scenario, to be transmitted to the others. Eventually, this group will be the closest one to the most altered area. But if these conditions are not met, the method will continue trying to find a solution.

## 5. SIMULATIONS AND DISCUSSION

### 5.1 Environment

We have tested our method in the $100m \times 100m$ environment depicted in the Fig.1(a), for a team formed by 7 agents. We fix the visual range of the agents to $30m$. The communication range between the agents is $30m$ as well, and there must be line-of-sight between the agents to establish connectivity. We set the velocity for all the agents to $2m/s$. The mission of the agents is to form a chain, if it is possible, from $x_{base} = [5,5]$ to $x_{goal} = [80,80]$.

We generate randomly the scenario variations, as well as the initial positions of the agents. The robots start from these positions with the initial map of the environment, Fig.1(a). All of them know the locations of the base station, the goal and the initial positions of the whole team, but not the variations. Three types of variations might occur: closures, appearance of small obstacles, and openings. We have run 100 trials, randomly producing 3 closures of doors, 5 new random obstacles and 5 openings of doors in each trial. Three reasons could prevent to obtain a solution: i) the base station and/or the goal are completely enclosed by obstacles; ii) there no exists a possible chain with the available agents, due to the new obstacles; or iii) some agents are trapped. Both situations are evaluated, with and without possible solution.

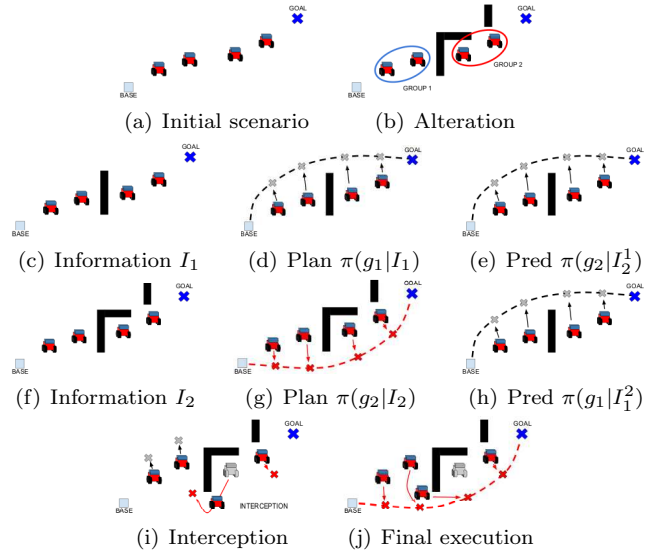

Fig. 2. Simplified explanation of the method

### 5.2 Comparisons

The proposed method is compared with two alternative solutions. The first one considers the ideal case in which all the agents have full knowledge of the new environment and are connected in the entire scenario, despite the obstacles. Therefore, if a solution exists, the method finds it and the agents directly form a chain.

The second one establishes a policy which guarantees the regrouping of all the agents. In Ulam and Arkin (2004) four behaviours are proposed for communication reestablishment between multi-robot teams. The four approaches guide the robots to: i) closest open space; ii) stored waypoints during the mission; iii) nearby inclines; iv) last known position of the nearest teammate. All these strategies depend on specific locations of the environment, thus we cannot consider them because, with random appearance of obstacles, some areas could be completely blocked. Therefore in our case, it is more appropriate to use some agents to reconnect the rest of the team. In Habibi and McLurkin (2014) a group of robots is used to extend a tree and guide the disconnected agents to a goal position, the root. This method does not consider that the root robot can be completely isolated, a possible case in our problem. Moreover, the number of agents in the team could not be enough to reach all the areas of the scenario, due to the new obstacles.

We consider that for the problem dealt with here, a searching strategy, with a searcher group, is the most proper policy to reconnect the entire team. The *searcher* group is selected to look for the rest of the agents, which wait at their initial positions until some searcher agent arrives. The selected searcher group is the closest, not trapped, group to the base station. One agent acts as leader and the rest follow him, within $c_{range}$. It does not have any knowledge of the new environment beyond its visual range, so that it discovers the changes during the seeking. When new agents are reconnected, the method chooses a new agent to reach and a new leader by proximity. The distance between agents is computed by a FMM method. The searching strategy acts in the same way as the prediction planner in case of detecting a trap situation, described in Sect.4.2. This way it assures the regrouping, if it is possible.

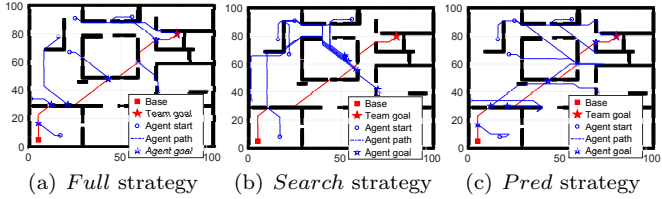

(a) *Full* strategy  (b) *Search* strategy  (c) *Pred* strategy

Fig. 3. Trajectories of the agents in the tested methods. In (b), only the paths until reconnecting the entire team are illustrated for clarity. After reconnection, the agents also form the chain.

The results of the proposed comparisons are denoted as *Full* when the agents have full information of the environment and connectivity, *Search* for the solution using the searching strategy, and *Pred* for our technique in which the robots only have partial information around their field of view and the information from the robots of its own group. The examples of these strategies are depicted in Fig.3. With *Full* in Fig.3(a), all the agents have the entire information of the changed environment and go directly to form a chain, depicted with red line. With *Search* strategy depicted in Fig.3(b), the agents are sequentially reconnected until form a unique group. Fig.3(c) depicts the trajectories of the agents with the proposed *Pred* approach. The agents are reconnected, forming the chain, after discovering new obstacles in the scenario.

### 5.3 Results

We measure the difficulty of the scenario by means of the number of the groups in which the team is split when the scenario changes. The evaluated metrics are: i) the time to reconnect the entire team, ii) the time to fulfill the mission (if a solution exists), and iii) the total distance travelled by the team. The reconnection time is the time to merge the robots in a unique group. The mission time is the time when the agent, at the goal location, is connected to the base station through its teammates. The mission time includes the reconnection time. The distance is the sum of the distances travelled by all the agents.

After running 100 random trials, we have obtained 48 scenarios with a possible solution and 52 where there is not possible to fulfill the mission forming a chain. The video of some simulated scenarios can be found in the link [1].

*Scenarios with solution.* The results for scenarios with solutions are depicted in Fig.4. In Fig.4(a), the number of groups are depicted. The first three bars correspond to the minimal, mean and maximum number of initial groups for the 48 scenarios. The other bars correspond to the maximum number of groups obtained by *Search* and *Pred* algorithms, respectively. As can be observed, with *Pred* algorithm the agents are separated into more groups in order to search other groups. While in *Search* routine, when some group is reconnected, their agents are following the searcher agent and never disconnected. So the values are exactly the same as for the initial groups. Finally, in both, *Search* and *Pred* strategies, the agents merge in a unique group.

Fig.4(b) depicts the minimal, mean and maximum times to reconnect the entire team for *Search* and *Pred* routines.

[1] http://robots.unizar.es/data/videos/iav19yamar.mp4

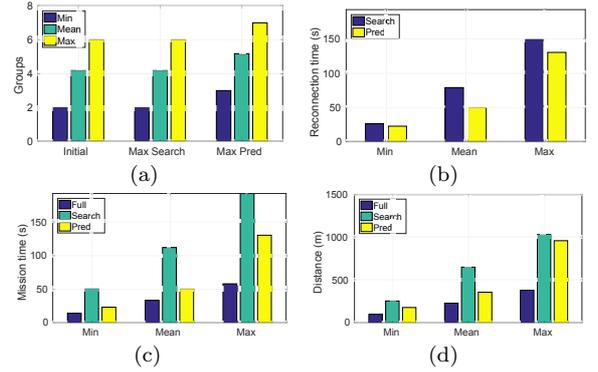

Fig. 4. Results of scenarios with possible solution. (a) Number of groups. (b) Time to reconnect the entire team. (c) Mission time. (d) Travelled distance.

Using predictions, all the agents are moving, attempting to reestablish communication with the rest of groups, so the reconnection occurs faster. *Full* routine is not included here because from the beginning all the agents have connectivity. Fig.4(c) shows the times to accomplish the mission (forming a chain) for the three routines. With *Full* routine the agents go directly to their new objectives to form a chain, so we use this time as a baseline for comparison. The values for the *Search* approach are much higher, because first all the groups recover the communication and, only then, the team forms the chain. In mean, it takes $111.9sec$ to fulfill the mission for *Search* strategy, against $32.91sec$ of *Full* routine. The proposed method *Pred*, fulfills the mission in $50.42sec$, in mean. The maximum values of times denote that there are some scenarios which are highly altered, so *Search* and *Pred* approaches require more time to discover the new obstacles and find the other groups of agents. The minimum and mean times of *Full* and *Pred* do not differ too much, because when the variation of the scenario does not affect to the initial plans of all the groups, the groups are reconnected with the first or few predictions, fulfilling the mission quickly. However, when the alterations are substantial, the groups require to travel more distance, discovering new obstacles, to find other groups. The worst scenario of *Pred* requires $130.3s$ against $57.66s$ of *Full*. As can be seen, the results of connectivity recovering and mission accomplishing times are exactly the same for *Pred*, because both tasks are accomplished simultaneously.

Fig.4(d) depicts the total distance travelled by all the team. In mean, using predictions, the agents travel less distance than using the searching strategy, $353.8m$ against $647.4m$. This is logical because in *Search* strategy the agents regroup and form a chain sequentially. Whilst, with the proposed *Pred*, the agents perform both tasks simultaneously. The same occurs if we observe the maximum distances, but the difference is not so significant, $956m$ for *Pred* against $1029m$ for *Search*. This happens because, in some scenarios with significant variations, there are many agents intercepting agents of other groups (Alg.3), because the latter ones have observed less variations. Once again the travelled distance for *Pred* is higher than for *Full*, since in the latter all the team has a full knowledge of the environment, and the agents go directly to the goals.

As can be observed, the deviation from *Full* strategy is more significant for *Search*. The mean travelled distance is 187% higher than for *Full*, against 57% of increase for the proposed *Pred*. In the same way, the increase of time is 240% for *Search* and 53% for *Pred*.

*Scenarios with no solution* The results for scenarios with no solution are represented in Fig.5. Since the absence of solution could be caused by a confinement of some agents by obstacles, we evaluate the groups at the end of the mission. Here, the mission finishes when the agents realize that they are not able to form a chain to transmit the information to the base station. Again, the first three bars correspond to the number of groups at the beginning of the mission.

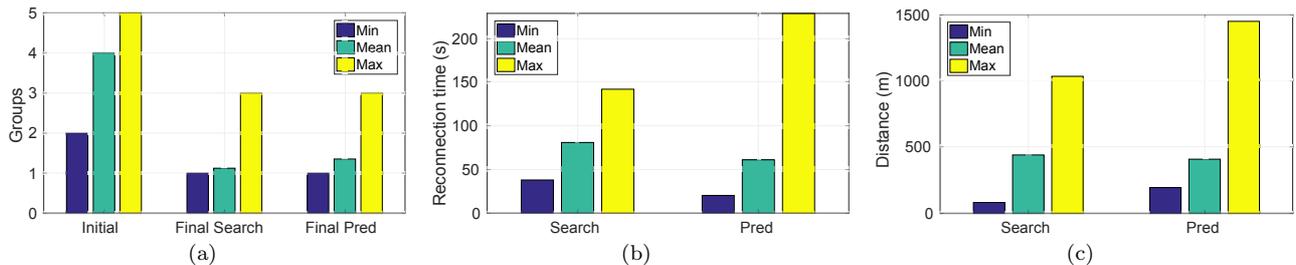

Fig. 5. Results of scenarios with no solution. (a) Number of groups. (b) Time to reconnect the entire team. (c) Travelled distance.

As we can observe, the minimum and maximum number of groups for $Search$ and $Pred$ routines are identical. The maximum values indicate that for the worst scenarios the proposed $Pred$ planner is able to determine that it is not possible to reconnect more groups, so the mission is unfeasible. The means of the groups are slightly worse for $Pred$ than for $Search$, 1.35 groups against 1.12 respectively. It occurs because $Pred$ has failed in four scenarios, due to a mismatch in goal allocation. Prediction planner, as opposed to searching routine, allows disconnections of the agents. So, when two agents of two different groups are allocated for going to search one to each other, they can be permanently exchanging their positions without merging in a unique group, appearing cyclic behaviours for these groups. This is also reflected in the maximum time and travelled distances. When this happens, it can be detected by the algorithm; a solution would be to fix the chain path planned by one of them, enforcing the other agent to move toward it, resolving the cyclic situations. In this kind of situations, our $Pred$ method would be more like the $Search$ method, reducing the maximum times for reaching the solution or for being aware that there is no solution. Despite this, the mean times of the proposed planner outperform the searching routine, $60.99sec$ against $80.84sec$ respectively. This corresponds to 33% of increase for the time. In the same fashion, the mean distance is higher, but only 8%.

## 6. CONCLUSIONS

In the present work we have presented a distributed method to coordinate a team of robots, in order to recover the connectivity after communication failures caused by unpredictable events, such as appearance of new obstacles or lost of line of sight. The proposed approach employs the predictions of actions of the different groups of agents to reconfigure the team.

With the proposed strategy, all the groups of disconnected agents seek for the rest of groups to restore the connectivity and also form the chain to reach the goal location. The method is compared against other two strategies: the ideal one in which all the agents have full connectivity and information of all the modifications of the scenario ($Full$); and another in which a group of agents is used to search the rest and reconnect them forming a unique group ($Search$). Our approach outperforms the searching strategy, since all the agents are moving seeking for the rest. Obviously, the results for our prediction-based planner are worse than for $Full$ strategy, because it spends a lot of time discovering new obstacles and new paths for the whole team in the environment. But if it exists, a solution is found.

In this paper we have focused on chain formations of the robots. In the future, we want to generalize the proposed method for different deployment strategies. For instance, we will consider the tree formation used in our previous work Marchukov and Montano (2017). We also plan to carry out real world experiments, in scenarios with human presence, which exhibit natural environment alterations from the point of view of the robots.